\documentclass[lettersize,journal]{IEEEtran}
\usepackage{amsmath,amsfonts}
\usepackage{algorithmic}
\usepackage{algorithm}
\usepackage{array}
\usepackage[caption=false,font=normalsize,labelfont=sf,textfont=sf]{subfig}
\usepackage{textcomp}
\usepackage{stfloats}
\usepackage{url}
\usepackage{verbatim}
\usepackage{graphicx}
\usepackage{cite}
\usepackage{xcolor}
\usepackage{enumitem} 
\usepackage{cancel}

\newif\ifshowcolor
\showcolorfalse  

\ifshowcolor
  \usepackage{xcolor}
\else
  \usepackage{xcolor} 
  \renewcommand{\textcolor}[2]{#2}
  \renewcommand{\color}[1]{}
  
\fi
\begin{document}

\title{Edge-Assisted Collaborative Fine-Tuning for Multi-User Personalized Artificial Intelligence Generated Content (AIGC)}
\author{Nan Li, Wanting Yang, Marie Siew, Zehui Xiong, Binbin Chen, Shiwen Mao, Kwok-Yan Lam}



\maketitle

\begin{abstract}
Diffusion models (DMs) have emerged as powerful tools for high-quality content generation, yet their intensive computational requirements for inference pose challenges for resource-constrained edge devices. 
Cloud-based solutions aid in computation but often fall short in addressing privacy risks, personalization efficiency, and communication costs in multi-user edge-AIGC scenarios. 
To bridge this gap, we first analyze existing edge-AIGC applications in personalized content synthesis, revealing their limitations in efficiency and scalability.
We then propose a novel cluster-aware hierarchical federated aggregation framework.
Based on parameter-efficient local fine-tuning via Low-Rank Adaptation (LoRA), the framework first clusters clients based on the similarity of their uploaded task requirements, followed by an intra-cluster aggregation for enhanced personalization at the server-side. Subsequently, an inter-cluster knowledge interaction paradigm is implemented to enable hybrid-style content generation across diverse clusters.
Building upon federated learning (FL) collaboration, our framework simultaneously trains personalized models for individual users at the devices and a shared global model enhanced with multiple LoRA adapters on the server, 
enabling efficient edge inference; 
\textcolor{red}{meanwhile, 
all prompts for clustering and inference are encoded prior to transmission, thereby further mitigating the risk of plaintext leakage.}
Our evaluations demonstrate that the framework achieves accelerated convergence while maintaining practical viability for scalable multi-user personalized AIGC services under edge constraints.


\end{abstract}


\section{Introduction}
\begin{figure*}[htbp]
	\center{\includegraphics[width=0.85\textwidth]{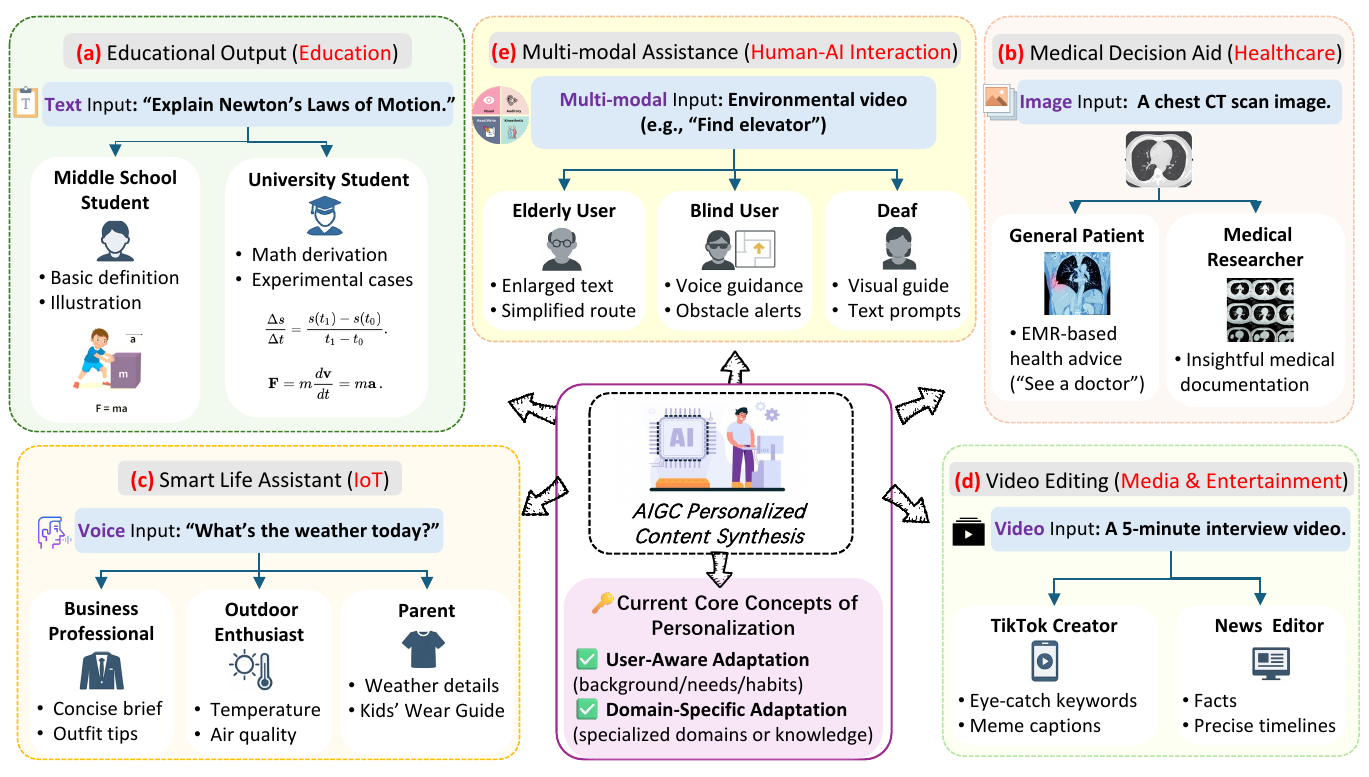}} 
	\caption{\textbf{Examplary applications of personalized Edge-AIGC models.} Adaptable foundation models enable multi-modal inputs (text, image, voice, video) to run on edge devices (e.g., smartphones, tablets, PCs), supporting applications across education, healthcare, IoT, entertainment, and human-AI interaction.}
	\label{applications}  
\end{figure*}

Artificial Intelligence Generated Content (AIGC) models, capable of multi-modal content generation (e.g., 
text-to-image, image-to-image, and text-to-video), have captured widespread attention due to their 
capabilities in synthesizing contextually intelligent content. 
For example, state-of-the-art image synthesis foundation models like DALL·E, Stable Diffusion Models (SDMs), and GPT-4o leverage advanced architectures to produce high-fidelity images from concise prompts.
Driven by real-world application demands, these generative models are shifting their focus from general-purpose tasks (e.g., image synthesis and style transfer) \cite{ahn2024dreamstyler, frenkel2024implicit} to client-side personalized content generation,
where client preferences are reflected in the outputs. Such AIGC-driven personalization is increasingly being applied across diverse fields, such as user-specific therapeutic recommendations in healthcare (Fig. \ref{applications}(b)) \cite{wang2025areas}, personalized life assistants in IoT (Fig. \ref{applications}(c)), or customized video editing services in entertainment (Fig. \ref{applications}(d)).

\begin{figure}[t]
    \centering
    \includegraphics[width=0.9\linewidth]{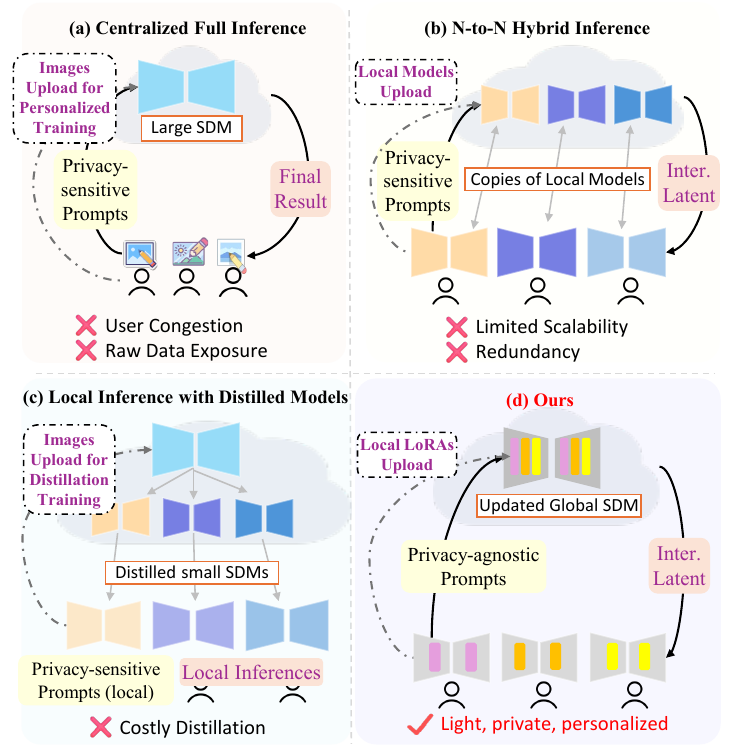}
    \caption{\textbf{Four key methods for accelerating DM inference locally.} Compared to the others, our proposed ``1-to-N Hybrid Inference" integrates multiple LoRAs into a global DM, enabling parallel multi-user inference via a shared latent—
    without requiring raw data upload or exposing sensitive prompts.}
    \label{hybridinfer}
\end{figure}

However, unlike Generative Adversarial Networks (GANs), which generate samples in a single forward pass, Diffusion Models (DMs) rely on a computationally intensive iterative denoising process (typically 100 to 1000 steps). 
This iterative nature, coupled with the considerable storage demands of AIGC models, poses challenges for deploying such systems on resource-limited end devices.
To mitigate these challenges, in the conventional setting (Fig.~\ref{hybridinfer}a), users typically upload their raw data along with the generation prompts to either powerful cloud servers or capable edge servers for full inference, which, however, raises concerns about raw data leakage.
With growing user expectations for low latency and privacy-preserving generation process, server-client hybrid inference frameworks are gaining attention in recent research \cite{du2023exploring,xie2025ec, yan2024hybrid, yang2025efficient}.

Hybrid inference frameworks aim to distribute computation between the server (for initial processing) and devices (for local denoising), based on task-specific prompts, as depicted in Fig. \ref{hybridinfer}(b-d). However, under a semi-honest server setting, 
current hybrid implementations incur $(i)$ a certain degree of \textbf{storage redundancy} and $(ii)$ \textbf{potential information leakage} stemming from user-submitted interaction data (e.g., hybrid inference queries). 
In particular, EC-Diff \cite{xie2025ec} explores cloud-side acceleration strategies for single-user per-model cases (Fig. \ref{hybridinfer}b), yet becomes storage-intensive when scaled to multi-user and requires uploading plaintext prompts that often contain identifiable user-preference terms (e.g., diagnostic terms like ‘\textit{dermatofibroma}’), creating server-side data exposure risks.
FedBip \cite{chen2025fedbip} also requires sharing domain- and instance-level plaintext prompts for server inference.
    Hybrid-SD \cite{yan2024hybrid} requires distilling a separate edge model per user (Fig. \ref{hybridinfer}c), causing costly computation on the cloud. 
    

Building upon this, our work focuses on reducing server-side storage without sacrificing personalization accuracy, \textcolor{red}{while enhancing user privacy by replacing plaintext prompts with encoded representations} in edge-AIGC systems.
Specifically, as depicted in Fig. \ref{hybridinfer}(d), our framework enables clients to fine-tune locally in a lightweight manner via Low-Rank Adaptation (LoRA) \cite{hu2022lora} federated fine-tuning, and a globally shared model is generated. By updating only a small subset of parameters while keeping the main model frozen, LoRA significantly reduces computational overhead.
Privacy is preserved through Federated Learning (FL) \cite{2023The}, 
which supports collaborative personalized training across multiple users by sharing only model updates and privacy-agnostic prompts with the server. 
Leveraging the collective knowledge of the trained model, a globally shared LoRA can be reused by local clients with similar downstream inference tasks, thereby reducing server-side storage redundancy.

\textcolor{red}{Under the semi-trusted server assumption, }we leverage DMs as the generative backbone and propose an FL-based fine-tuning framework for collaborative edge personalization. The key contributions are as follows: 

\begin{itemize}
    \item We first systematically analyze the potential application scenarios for personalized content synthesis in edge-AIGC systems under multi-user environments, while highlighting the limitations of existing approaches. 
    \textcolor{red}{Specifically, we investigate two key challenges in
    $(i)$ the performance limitations of standard FL/PFL under \textbf{increasing data heterogeneity} and \textbf{larger AIGC model architectures}, and $(ii)$ the \textbf{protection of user preferences} contained in user-submitted prompts (Section II).}

    \item We then present our cluster-aware hierarchical federated aggregation framework for scalable multi-user edge personalization. It combines intra-cluster aggregation for enhanced personalized feature representation and inter-cluster aggregation to integrate diverse personalization into a general global model for hybrid inference. Unlike conventional clustering, \textcolor{red}{we avoid raw data transmission by using embeddings, and employ textual inversion-based tokens for hybrid inference on the server} (Section III). 
    \item Experiments on real-world data confirm that our multi-style LoRA-enhanced global model shows around 40\% better latent space alignment (best-case 0.6× Fréchet Inception Distance (FID) on sketch generation) than the baseline. It successfully serves multiple devices for parallel personalized generation, reducing edge-side computation while preserving style precision (Section IV).
\end{itemize}

\section{Applications and Challenges in FL-based Edge-AIGC} \label{section2}

In this section, we present potential personalized application scenarios for edge-AIGC architectures and the key challenges that must be addressed for real-world deployment.

\subsection{Applications of Personalized Edge-AIGC} \label{section2.A}
Figure \ref{applications}(a–e) demonstrates five practical cases of personalized content generation across different areas on edge devices.
We categorize the personalization into three types based on their optimization focus: $(1)$\textit{ client-aware personalization}—adapts outputs to individual user characteristics, $(2)$ \textit{domain-specific personalization}—tailors outputs to specialized knowledge domains, and $(3)$ \textit{modal-aware personalization}—optimizes for cross-modal interaction patterns.

\subsubsection{Client-Aware Personalization}
This case highlights that output should adapt to user attributes (e.g., age or preferences).
Fig. \ref{applications}(d) illustrates this concept in media content creation when generating a news poster from a 5-minute interview video. For news editors, a data-driven summary is enriched with precise details like timelines, 
whereas social media creators prefer a more eye-catching version. 
Textual Inversion \cite{gal2022image} and DreamBooth \cite{ruiz2023dreambooth} enable effective few-shot (3—5 images) adaptation for such scenarios, but their parameter-isolated learning paradigm—independently optimized per-user embeddings
—poses a barrier for dynamic edge deployment.

\subsubsection{Domain-Specific Personalization}
Domain-aware personalization meets broader field-specific demands. 
As shown in Fig. \ref{applications}, (a) educational materials favor colorful cartoon visuals, 
whereas (b) medical imaging is characterized by precise 3D analysis.
Data across fields exhibit fundamental divergences in both data characteristics (artistic stylization vs. clinical realism) and structures 
(2D sequential layouts vs. 3D spatial organizations), making cross-domain adaptation significantly harder than personalizing for individual users. 
Managing such feature heterogeneity becomes critical for effective FL-based multi-domain personalization.

\subsubsection{Modal-Aware Personalization}
Beyond content-level (i.e., client-aware and domain-specific) personalization, modal personalization represents a critical and rapidly growing area in AIGC evolution. It enables models to dynamically select input/output modalities (e.g., text/voice/image) based on user profiles.
Consider the intelligent personal assistant scenario in Fig. \ref{applications}(e): visually impaired users typically prefer audio-based outputs, whereas hearing-impaired individuals are more likely to rely on text-based modalities.
The integration of multi-modal capabilities has enabled more flexible interaction, further expanding the range of real-world application scenarios.

\subsection{Challenges of Developing FL-based Edge-AIGC} \label{Challenges}
While integrating FL into edge-AIGC frameworks offers promising benefits—particularly in the preservation of raw data while supporting multi-user coordination—it also poses several technical challenges. 

\begin{itemize}

\begin{figure}[t]
    \centering
    \includegraphics[width=0.8\linewidth]{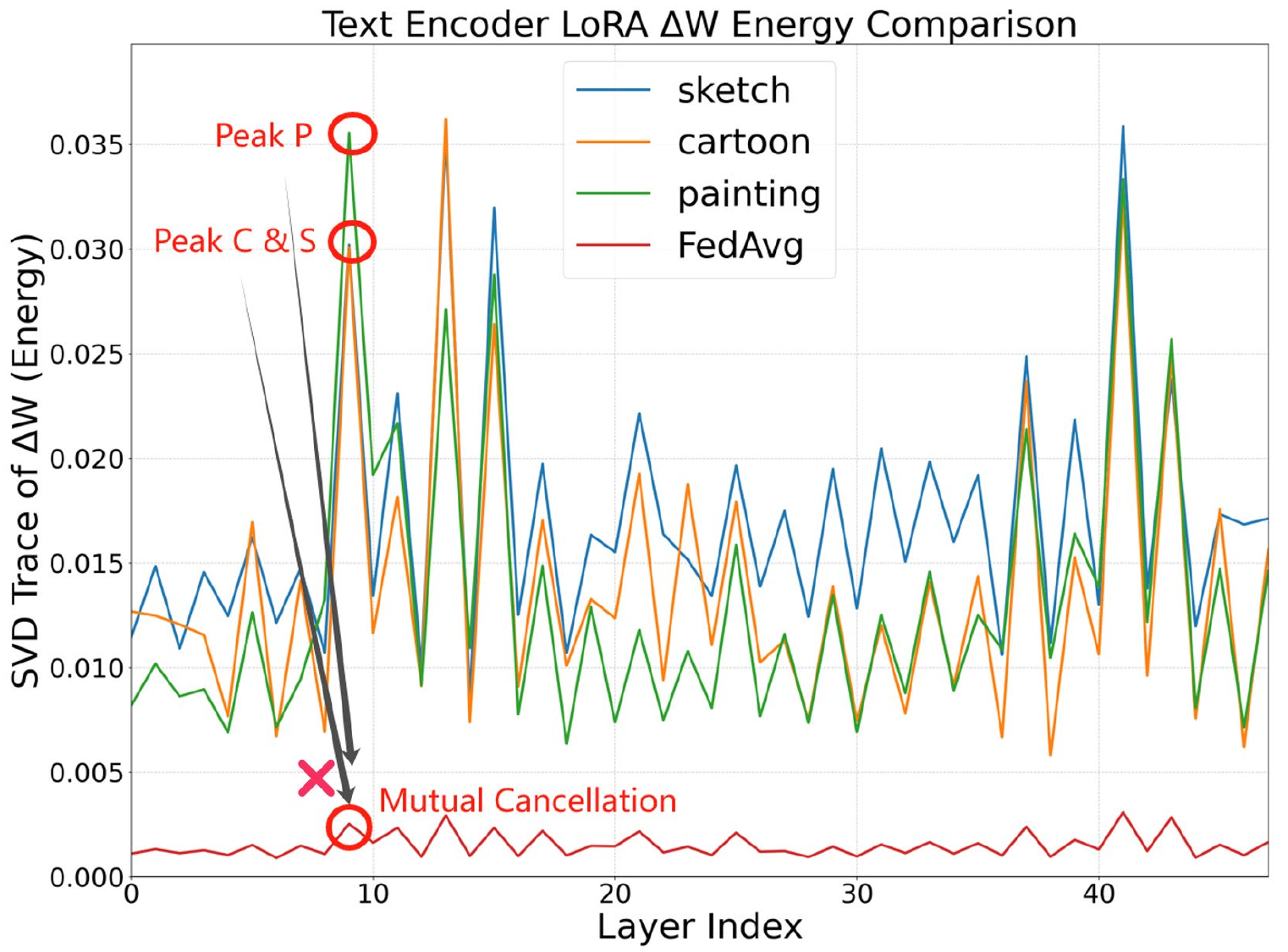}
    \caption{\textbf{Feature energy (SVD) of LoRAs vs. FedAvg by layer.} Per-layer SVD trace (sum of squared singular values, style representation energy) reveals FedAvg's personalization loss from conflicting client updates.
    }
    \label{fedavg}
\end{figure}

    \item \textbf{Dual Heterogeneity: Feature Distribution and Device Resource Availability}. \label{dulahete}
    When targeting multi-user personalization in FL-based edge-AIGC systems, heterogeneity extends beyond traditional FL's label heterogeneity assumptions. Two critical dimensions are pronounced: (1) \textit{Feature-level heterogeneity}—as noted in Section~\ref{section2.A}, even identically labeled inputs can exhibit varying 
    feature distributions, creating complex non-IID effects; (2) \textit{Device-driven 
    heterogeneity}—hardware constraints force LoRA rank adaptation: high-capacity devices employ higher ranks (64/128) for richer parameterization, while resource-limited ones use lower ranks (4/8/16) to conserve resources. This dual heterogeneity invalidates standard aggregation methods \cite{wang2024flora}: for example, the widely used FedAvg algorithm fails to properly reconcile divergent LoRA matrices without specialized adaptation, as shown in Fig. \ref{fedavg}.
    Tackling multidimensional heterogeneity beyond standard FL assumptions is critical to multi-user personalization in edge-AIGC.

     \item \textbf{Classic PFL Methods Fail in AIGC}. 
Personalized FL (PFL) frameworks enhance standard FL by supporting client-specific personalization, aligning with our goal.
Common PFL approaches \cite{tan2022towards} 
    include adding a regularization term for the local objective, performing additional fine-tuning on each client, or decoupling parameters into globally shared and client-specific components.
    While these approaches have shown success with conventional neural networks (e.g., CNNs), they face challenges when applied to AIGC architectures due to fundamental differences in model complexity and scale.
    For example, Phoenix \cite{stanley2024phoenix} adopts parameter decoupling by assigning the U-Net’s final decoder block for client-specific adaptation while keeping the other layers shared in the global model. 
    However, this design overlooks the critical role of attention layers in capturing fine-grained features \cite{frenkel2024implicit}, limiting its effectiveness for nuanced personalization.

    \item \textbf{Personalization-Driven Privacy Risks in FL-based Edge-AIGC}.\label{privacyrisk}
    \textcolor{red}{In FL-based edge-AIGC systems, directly transmitting plaintext prompts to semi-trusted servers can pose privacy risks—particularly attribute leakage through client clustering in training and user preferences in hybrid inference requests.}
    Specifically, $(i)$ strong heterogeneity across users (as mentioned in ``\textit{Dual Heterogeneity}”) often necessitates additional personalization strategies like client clustering, which groups similar clients based on auxiliary prompts (e.g., local data statistics), but \textcolor{red}{these user-specific prompts may disclose private attributes (e.g., style preferences or medical terms like ‘\textit{dermatofibroma}’), potentially enabling adversaries to perform property inference or related attacks.} 
    In addition, $(ii)$ 
    when the server manages both training and inference processes, its access to local updates (e.g., LoRA adapters) and the pre-trained backbone enables potential inversion attacks—the server may reconstruct private data features or latent attributes from model outputs.
    Given these risks, it is essential to reduce the exposure of plaintext during both training and inference, in order to mitigate potential privacy leakage in semi-trusted FL-based edge-AIGC systems.

\end{itemize}
\section{Federated Multi-User Personalization} \label{section_solution}
\begin{figure*}[htbp]
	\center{\includegraphics[width=16cm]{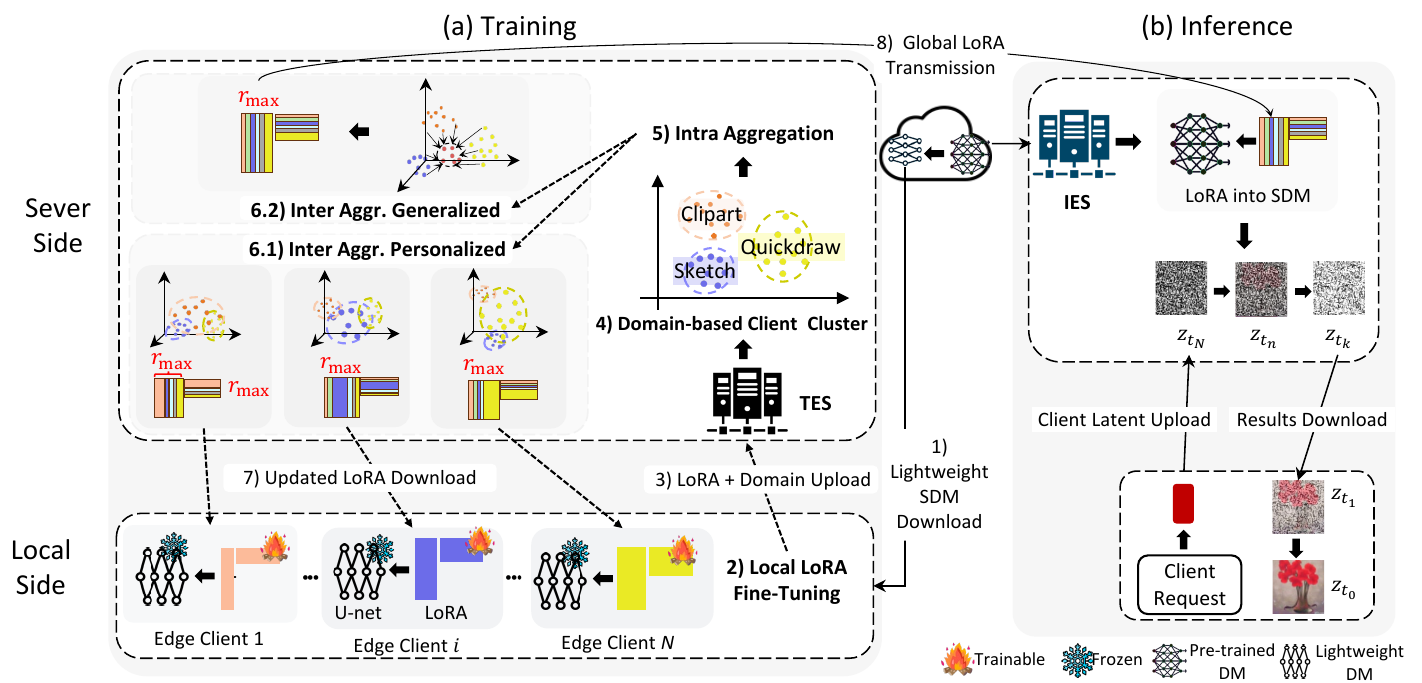}} 
	\caption{\textbf{A federated framework for fine-tuning personalized diffusion models across edge devices.} Clients are grouped by domain (e.g., image style) based on their dataset characteristics. 
    After cluster-level aggregation, domain-specific personalized models are sent back to each group, while a global LoRA model is delivered to the IES for hybrid inference—balancing personalization with improved generalization over the base SDM, \textcolor{red}{for future requests.}}
    \label{wholemodel}  
\end{figure*}


Following the key challenges for multi-user personalization 
in Section \ref{section2}, Fig. \ref{wholemodel} presents our solutions, consisting of two components: $(1)$ a \textbf{\textit{cluster-aware hierarchical}} FL training system 
(Fig. \ref{wholemodel}a), and $(2)$ an \textbf{\textit{optimized hybrid inference}} pipeline (Fig. \ref{wholemodel}b).
During the training process, after local fine-tuning, the Training Edge Server (TES) performs client clustering based on user preferences (e.g., sketch-style versus cartoon-style), followed by intra-cluster aggregation for enhanced personalization and inter-cluster aggregation for generating mixed-style LoRA adapters.
During hybrid inference, the shared global LoRA adapter works together with a pre-trained base model on the Inference Edge Server (IES) to perform denoising.
Compared to the baseline approaches in Fig. \ref{hybridinfer}, our framework greatly enhances system scalability, allowing cluster-aware aggregated models to be efficiently shared among local users and reused in hybrid inference.

\subsection{Overview: Privacy Considerations in Our System Design}

\textcolor{red}{
In our system, we adopt the practical assumption that servers are semi-trusted—i.e., they adhere to protocols but may passively analyze received information. 
Motivated by the privacy vulnerability consideration in Section \ref{privacyrisk}, we decouple server roles into TES (training) and IES (inference) to limit sensitive data access, assuming no collusion between them.} 
Specifically, the TES is responsible for LoRA parameter 
aggregation and is restricted from accessing the full base model architecture. 
The IES holds access to the pre-trained base model for hybrid inference, and retains a generalized global LoRA, which can only produce style-neutral intermediate representations, thus avoiding leaking user-specific features to the server. 
Moreover, to prevent privacy leakage from client-submitted auxiliary data  (i.e., domain keywords or inference prompts), \textcolor{red}{we replace plaintext transmission of such data with encoded representations.} For example, a domain term ‘\textit{sketch-style}’ is transformed into obfuscated tokens ‘\textit{\textless user-specific\textgreater}’, which are semantically opaque to servers.

\subsection{Cluster-aware Hierarchical Multi-user Co-training}
The co-training framework focuses on: 
(a) client-side LoRA fine-tuning with capability-adaptive LoRA ranks (\textit{Step 2 in Fig. \ref{wholemodel}}), and (b) server-side cluster-aware hierarchical LoRA aggregations (\textit{Step 4-6}). 
Three sequential phases are executed: 

\textbf{Initialization and Local LoRA Fine-Tuning.}
 Built on DMs, our system applies LoRA to 
 the attention layers of the text encoder and U-Net modules—where self-attention layers capture stylistic nuances and cross-attention layers align outputs with text prompts.
For client inputs with minimal semantic gaps from the system’s text encoder, we suggest fine-tuning only self-attention layers. 
 Client data is structured as $(image, domain, label)$ triplets, where $domain$ indicates client-specific styles 
 (e.g., sketch style) to guide clustering. 
 To minimize communication overhead while preserving personalization, \textit{only low-rank LoRA updates and privacy-agnostic style descriptors for clustering are transmitted to the server.}

\textbf{Domain-based Client Clustering and Intra-cluster Aggregation.}
    In contrast to traditional clustering methods relying on sensitive user-uploaded statistics (e.g., 
    data distributions for K-means++), %
    our approach prevents the semi-trusted edge server from accessing users' plaintext domain information.
    Clients upload only prompt embeddings (e.g., images or text embeddings) for similarity computation
    and requires no predefined cluster counts. 
    To aid comprehension, we present the domain expressions in plaintext format in Fig. \ref{wholemodel}.
    Within each cluster, LoRAs are aggregated via weighted averaging to enhance domain-specific features through member consistency, though using it across clusters may cause feature cancellation due to multi-style mixing (Fig. \ref{fedavg}). 
    To address rank heterogeneity, we introduce \textit{dynamic median-aligned padding}: (1) compute the cluster median rank \( r_{\text{median}} \), (2) truncate higher-rank matrices, and zero-pad lower-rank ones. It mitigates alignment noise better than fixed \( r_{\text{max}} \)-padding. 

\textbf{Inter-cluster Aggregation.}
    Following the intra-cluster aggregation phase, 
    we implement inter-cluster aggregation to enhance cross-domain knowledge transfer and enable controlled style mixture via adjustable coefficients. 
    Inspired by Flora \cite{wang2024flora}, we preserve domain-specific features via stacking strategy, 
    then leverage dual similarity metrics—\textit{Domain Embedding Distance (DED)} and \textit{SVD Normalized Trace (SNT)}—to determine aggregation coefficients, 
    precisely controlling each domain’s contribution to the final output. 
    Specifically, \textit{DED} helps filter out semantically unrelated domains in the CLIP space; 
    \textit{SNT} measures structural similarity by quantifying layer-wise feature importance—closer \textit{SNT} values indicate more compatible LoRA representations.  
    \textit{DED} and \textit{SNT} jointly guide coefficient generation for controlled mixed-style aggregation.
    By the end of this phase, our co-training framework produces two distinct outcomes:
    (1) personalized models that capture each user's preferred style mixture (Fig. \ref{wholemodel} step 6.1); 
    (2) a shared global LoRA module that generates style-neutral representations (Fig.\ref{wholemodel} step 6.2). 

\begin{figure*}[t]
	\center{\includegraphics[width=15cm]{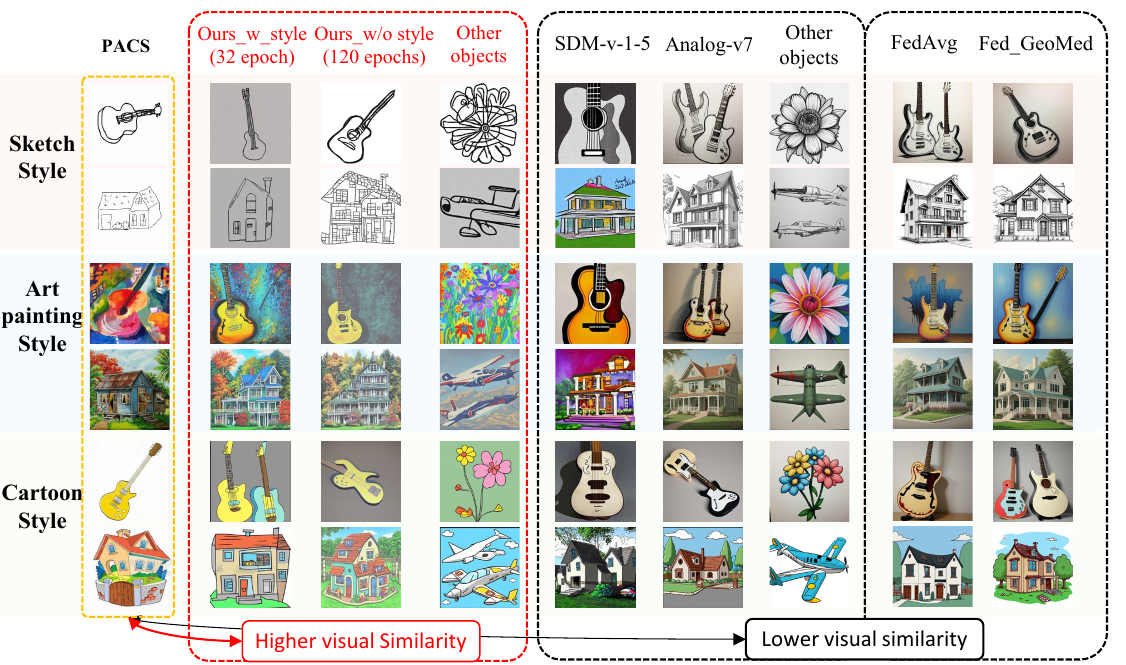}} 
	\caption{\textbf{Comparison results on the PACS dataset}: $(i)$ Without fine-tuning, two pre-trained SDMs (\textit{SDM-v1-5} and \textit{Analog-v7}) fail to match PACS’s styles; 
    $(ii)$ After fine-tuning and applying typical FL aggregation methods (i.e., \textit{FedAvg} and \textit{Fed\_GeoMed}), the global outputs show limited feature alignment.
    $(iii)$ Our outputs (\textit{``Ours\_w\_style"} and \textit{``Ours\_w/o\_style"}) better align with PACS’s styles, which with ``\textit{A \{label\} with \textless user-specific\textgreater style}" and ``\textit{A \{label\} with S/A/C style}" as input prompts separately.
    $(iv)$ Our method achieves better alignment on unseen objects (\textit{``flower", ``plane"}) that are absent in training data.
    }

	\label{loraagg}  
\end{figure*}

\subsection{Hybrid Inference Architecture}
Fig. \ref{case2} illustrates how the IES handles multi-user generation requests using a LoRA-enhanced global SDM. With a shared generic prompt (e.g., ``\textit{A dog}"), it first produces a semantically rich intermediate latent representation, embedding multiple style potentials. Each client device then applies its own local style condition prompts (e.g., $Client_1$: ``\textit{sketch style}"; $Client_2$: \textit{``art painting style}") to generate the final personalized output—all without sending private data off the device.
To activate the global LoRA module, 
server-side prompts must include style-specific tokens alongside the generic content prompt.
In our framework, to protect style attributes from exposure to the semi-trusted server, the textual inversion \cite{gal2022image} method is employed to replace explicit style words (e.g., ``$sketch$”) with private implicit tokens (e.g., ``$<user$-$specific>$"). 
This approach ensures reliable activation of the global LoRA while safeguarding original style information. Overall, by decoupling content and style at the feature level and aligning shared features on the server, our architecture achieves an optimal balance of system efficiency, customization flexibility, and privacy protection.


\section{Case study} \label{casestudy}

\subsection{Models, Datasets and Experiment Settings.}

\begin{figure*}[t]
	\center{\includegraphics[width=18cm]{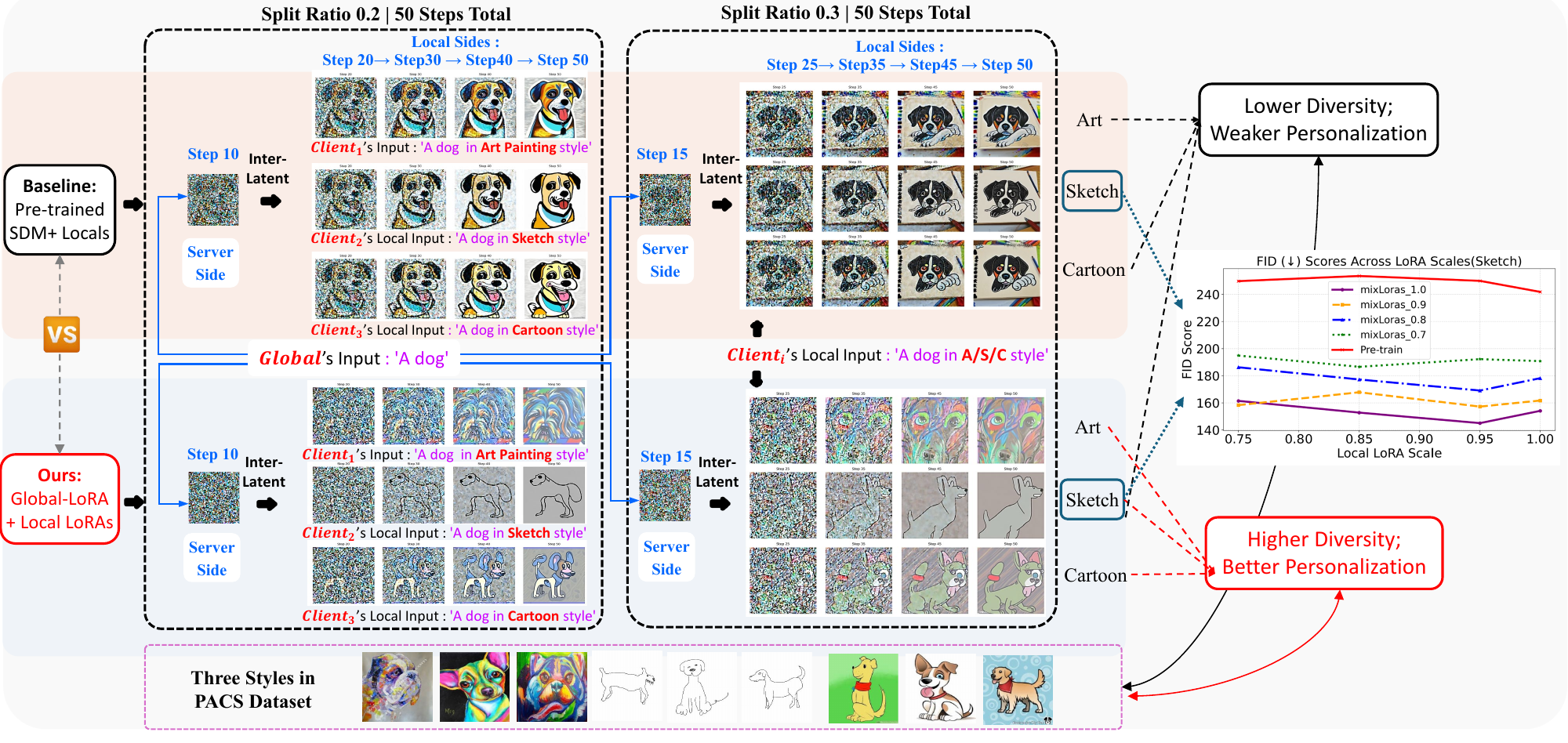}} 
	\caption{\textbf{Hybrid inference comparison: LoRA-enhanced SDM vs. pre-trained SDM.}
    Given a generic prompt (i.e., ``\texttt{\textit{A dog}}"), the server performs early-stage inference for 10 (0.2) or 15 (0.3) of the 50 total steps.
    Based on the same shared intermediate latent, 
    each client applies its own prompt 
    for downstream generation. Ours preserves both diversity and user-specific style features, whereas the vanilla pre-trained model exhibits weaker diversity and lacks personalization.
    As a representative case, the same result is validated by FID (Fréchet Inception Distance) scores of the $sketch$ style.}
	\label{case2}  
\end{figure*}

We evaluate two SDMs (\textit{SDM-v1.5} and \textit{Analog-v7})
as our AIGC base models to assess their fine-tuning potential. 
The PACS dataset (7 classes across 4 domains: Photo/Art/Cartoon/Sketch) is used for personalized image synthesis.
To intuitively assess federated fine-tuning performance, we simplify the whole process with rank-16 LoRA for all clients, 
bypassing intra-cluster entirely and directly performing the inter-cluster aggregation across three LoRAs; the FL simulation follows a zero-shot setting, where each unique style is assigned to a single user (with 100 images each). 
This setup ensures lightweight adaptation, updating only 1.18M parameters in the text encoder and 3.19M in the U-Net per client.
All experiments are conducted on NVIDIA GeForce RTX 4090 GPUs (24GB each) running CUDA 12.6.

\subsection{Baselines and Experiment Results.}  

\textbf{Style Alignment in Personalized Image Generation}: 
As shown in Fig. \ref{loraagg}, 
when given the same input prompts—``\textit{A \{label\} with \{sketch/art/cartoon \} style}"—the two pre-trained DMs (e.g., \textit{SDM-V-1-5} and \textit{Analog-v7}) show noticeable style deviations from the PACS. This discrepancy is primarily attributed to the models' varying interpretations of certain terms, particularly the semantic ambiguity of ``\textit{sketch}", but our models can effectively correct these deviations, generating images that not only better match the PACS style but also preserve diversity.
We also compare two classic FL aggregation strategies—\textit{FedAvg} reconstructs the global model via layer-wise weighted accumulation of client updates; \textit{FedGeoMed} applies geometric medians at each layer to enhance robustness against outlier gradients.
As reflected in Fig. \ref{fedavg}, due to the inconsistency in the direction of feature representations among local LoRAs, important personalized features tend to cancel out during aggregation, leading to suboptimal results.

In comparison, our fine-tuned variants (\textit{Ours\_w\_style} and \textit{Ours\_w/o\_style}, both using stack-based aggregation) produce more stylistically aligned results than the four baselines.
Specifically, \textit{Ours\_w/o\_style} replaces explicit style terms (e.g., ``\textit{sketch}") with implicit descriptors (e.g., \textit{\textless user-specific\textgreater}) in the prompts.
This setting allows for more privacy-aware control over style, but requires longer fine-tuning epochs. As shown in Fig. \ref{loraagg}, with a batch size of 2 and 100 local images, it takes 120 epochs to converge—nearly four times longer than \textit{Ours\_w\_style} (32 epochs).


\textbf{Cluster-based Aggregation for Personalized Outputs}: 
    Fig. \ref{case2} compares hybrid inference results between the pre-trained SDM ($Baseline$) and our LoRA-enhanced SDM ($Ours$) based on generating intermediate latent representations on the server. 
    With a total of 50 inference steps, the server conducts early-stage inference on 20\% (10 steps) or 30\% (15 steps) of the steps; under these split ratios, our outputs better align with the PACS test data distribution, particularly in the $sketch$ style.
    
    Quantitative evaluation using Fréchet Inception Distance (FID) (averaged over 100 images) confirms these observations, as seen in Fig. \ref{case2}—lower scores indicate closer statistical alignment with real images. 
    The $mixLoras \in [0.7,1.0]$ controls the contribution of the global LoRA to the outputs' stylization, while the local LoRA scale (0.75-0.95) adjusts the strength of client-side personalization; higher scale values indicate stronger personalization effects.
    Our method achieves optimal performance (FID=145) at $mixLoras$=1.0 with local scale=0.95, surpassing the pre-trained baseline (FID=240) by 40\% across all tested $mixLoras$ ranges (0.7-1.0).
    This demonstrates that the fine-tuned global model can enhance output quality compared to relying solely on localized adjustments.

\section{Future research directions}
To fully realize the potential of FL-powered Edge-AIGC systems, several critical open problems are considered:

\begin{itemize}
    

    
    \item \textbf{Maintaining Personalization under Dynamic Client Arrivals.} 
    Real-world edge AIGC applications involve dynamic user arrivals and diverse stylistic demands. New arrivals may possess distinct distributions and generation requirements compared to existing clients, calling for more adaptive solutions that do not rely on coarse-grained clustering when initializing new users.
    A promising future direction is to develop redundancy-aware adaptation strategies that determine whether new clients should reuse or align with existing aggregated LoRAs, or set up new clusters.
    Addressing this requires novel methods for quantifying 
    the similarity between a new client request and existing clients and LoRA models. 
    


    \item \textbf{Network-Aware Aggregation for Federated AIGC.} 
    In 
    real-world federated AIGC systems, clients face heterogeneous network conditions and device constraints. Some contribute updates less frequently due to limited bandwidth or power, while others provide more frequent or semantically richer updates. This imbalance can introduce aggregation bias and weaken personalization in the global LoRA model.
    Future directions include developing network-aware aggregation methods that incorporate both the communication capabilities of clients and the semantic relevance of their updates, such as adaptive weighting schemes or bandwidth-aware optimization.

    \item \textbf{Adversarial Attacks in Federated Edge-AIGC.} 
    The distributed nature of FL-based training makes it vulnerable to adversarial behaviors, such as data inversion and model poisoning by malicious clients.
    By injecting adversarially crafted data or uploading manipulated models (e.g., LoRA parameters), these users can influence
    the aggregation process, potentially 
    degrading the performance of downstream generation tasks.
    Therefore, robust detection and defense mechanisms are crucial for safeguarding the integrity of federated AIGC systems.
\end{itemize}

\section{Conclusion}
AIGC deployments are undergoing a shift from cloud-centric generation
to edge-driven adaptations.
To support this evolution, we have focused on personalized AIGC at the edge and 
have proposed a federated multi-user fine-tuning framework.
Based on LoRA-enhanced diffusion models, our approach has supported scalable and privacy-preserving content generation across 
\textcolor{red}{heterogeneous} devices and user needs.
Specifically, our cluster-aware hierarchical optimization framework has integrated federated learning with adaptive multi-LoRA to overcome three key limitations: server-side redundancy, high heterogeneity, and the failure of classic PFL methods in more complex and larger AIGC networks. 
Case studies have shown that our method better balances personalized inference accuracy and system scalability while preserving data privacy, advancing practical edge deployment of personalized AIGC. 





\bibliographystyle{IEEEtran}
\bibliography{IEEERef}

\section*{Biographies}  
\vspace{-3em}  
\begin{IEEEbiographynophoto}{NAN LI}
(nan\_li@mymail.sutd.edu.sg) is currently a Ph.D. student in the Information Systems Technology and Design (ISTD) Pillar at the Singapore University of Technology and Design (SUTD). Her research interests include cloud data security, federated learning, edge intelligence, and generative AI.

\end{IEEEbiographynophoto}

\vspace{-3em}  

\begin{IEEEbiographynophoto}{WANTING YANG}
(wanting\_yang@sutd.edu.sg) is currently a Research Scientist at the SUTD Wireless Innovation Centre (SWIC), serving as a Principal Investigator on a generative semantic communication project. Her research interests include semantic communication, deep reinforcement learning, martingale theory, edge computing, edge intelligence, and generative AI.
\end{IEEEbiographynophoto}

\vspace{-3em}  

\begin{IEEEbiographynophoto}{MARIE SIEW}
(marie\_siew@sutd.edu.sg) is a Faculty Early Career Award Fellow in the ISTD Pillar at the SUTD. 
She has received the Best Poster Awards at IEEE ICDCS 2024 and ACM/IEEE IPSN 2023.
Her research interests include edge computing, edge intelligence, network optimization, and federated learning.
\end{IEEEbiographynophoto}

\vspace{-3em}  

\begin{IEEEbiographynophoto}{ZEHUI XIONG}
(z.xiong@qub.ac.uk) is a Full Professor at Queen’s University Belfast, UK. A Highly Cited Researcher, he has published over 300 peer-reviewed papers in leading journals, with numerous Best Paper Awards from international flagship conferences. His research covers wireless networks, edge intelligence, semantic communications, generative AI, and the Metaverse.
\end{IEEEbiographynophoto}

\vspace{-3em}  
\begin{IEEEbiographynophoto}{BINBIN CHEN}
(binbin\_chen@sutd.edu.sg) is an Associate Professor in the ISTD Pillar at SUTD and serves as Deputy Director of Singapore’s Future Communications Research and Development Programme (FCP). 
He was a recipient of the Best Paper Awards in SIGCOMM’10. 
 His research focuses on wireless networking, distributed systems, and cybersecurity for infrastructures.
\end{IEEEbiographynophoto}
\vspace{-3em}  
\begin{IEEEbiographynophoto}{SHIWEN MAO}
(smao@auburn.edu) is a Professor and Earle C. Williams Eminent Scholar, and Director of the Wireless Engineering Research and Education Center at Auburn University. 
He is a co-recipient of Best Paper/Demo Awards of 12 conferences.
His research interests include wireless networks, multimedia communications, and smart grid.
\end{IEEEbiographynophoto}
\vspace{-3em}  
\begin{IEEEbiographynophoto}{KWOK-YAN LAM}
(kwokyan.lam@ntu.edu.sg) is the Associate Vice President (Strategy and Partnerships) and Professor in the College of Computing and Data Science at the Nanyang Technological University (NTU), Singapore. He is currently also the Executive Director of the Digital Trust Centre Singapore (DTC)
His research focuses on distributed systems, IoT and cyber-physical security, blockchain, biometric cryptography, and cybersecurity.

\end{IEEEbiographynophoto}

\vfill

\end{document}